\documentclass[letterpaper, 10 pt, conference]{ieeeconf} 

\IEEEoverridecommandlockouts 

\usepackage{flushend}
\usepackage{balance} %

\usepackage[mathlines,switch]{lineno}

\usepackage{array}
\usepackage[caption=false,font=normalsize,labelfont=sf,textfont=sf]{subfig}
\usepackage{textcomp}
\usepackage{stfloats}
\usepackage{url}
\usepackage{verbatim}
\usepackage{graphicx}

\usepackage[font=small,labelfont=bf]{caption}
\usepackage{multicol}
\usepackage[bookmarks=true]{hyperref}

\usepackage{booktabs, tabularx,dcolumn, float, multirow}
\newcommand*{\ditto}{\texttt{"}}

\DeclareMathVersion{nxbold}
\SetSymbolFont{operators}{nxbold}{OT1}{cmr} {b}{n}
\SetSymbolFont{letters}  {nxbold}{OML}{cmm} {b}{it}
\SetSymbolFont{symbols}  {nxbold}{OMS}{cmsy}{b}{n}

\makeatletter
\newcolumntype{d}{D{.}{.}{-1} } 
\newcolumntype{B}[3]{>{\boldmath\DC@{#1}{#2}{#3} }c<{\DC@end} }
\newcolumntype{Z}[3]{>{\mathversion{nxbold}\DC@{#1}{#2}{#3} }c<{\DC@end} }

\makeatother
\newcommand{\bd}[1]{\textbf{#1}}

\newcommand{\colrot}[2]{\parbox[t]{2mm}{\multirow{#1}{*}{\rotatebox[origin=c]{90}{#2}}}}

\newcommand\mc[1]{\multicolumn{1}{c}{#1}}

\newcommand{\secref}[1]{Sec.~\ref{#1}}
\newcommand{\equref}[1]{Eq.~\ref{#1}}
\newcommand{\figref}[1]{Fig.~\ref{#1}}
\newcommand{\tabref}[1]{Tab.~\ref{#1}}

\usepackage[noadjust]{cite}

\usepackage{amssymb}  
\usepackage{amsmath}
\DeclareMathOperator*{\argmax}{arg\,max}

\begin{document}

\title{\LARGE \bf Fast Explicit-Input Assistance for Teleoperation in Clutter}

\author{Nick Walker$^{1}$, Xuning Yang$^2$, Animesh Garg$^{2,3}$, 
Maya Cakmak$^1$, Dieter Fox$^{1,2}$, Claudia P\'{e}rez-D'Arpino$^2$%
\thanks{$^{*}$ This work was done during an internship at NVIDIA.}
\thanks{$^1$ University of Washington, $^2$ NVIDIA. $^3$ Georgia Institute of Technology. \href{mailto:nswalker@cs.uw.edu}{\texttt{nswalker@cs.uw.edu}}}%
}

\maketitle
\thispagestyle{empty}
\pagestyle{empty}

\begin{abstract}
The performance of prediction-based assistance for robot teleoperation degrades in unseen or goal-rich environments due to incorrect or quickly-changing intent inferences. 
Poor predictions can confuse operators or cause them to change their control input to implicitly signal their goal.
We present a new assistance interface for robotic manipulation where an operator can explicitly communicate a manipulation goal by pointing the end-effector. The pointing target specifies a region for local pose generation and optimization, providing interactive control over grasp and placement pose candidates.
We evaluate this explicit pointing interface against an implicit inference-based assistance scheme and an unassisted control condition in a within-subjects user study (N=20), where participants teleoperate a simulated robot to complete a multi-step singulation and stacking task in cluttered environments.
We find that operators prefer the explicit interface, experience fewer pick failures and report lower cognitive workload. Our code is available at: \href{https://github.com/NVlabs/fast-explicit-teleop}{\texttt{github.com/NVlabs/fast-explicit-teleop}}.
\end{abstract}

\IEEEpeerreviewmaketitle

\section{Introduction}

Robot telemanipulation is widely useful but demanding, even for skilled operators.
Acting in the world through a foreign embodiment with limited perception requires the user to reason not only about the task at hand but about the abilities and limitations of the robot, as well as the state of the environment.
Assistive teleoperation interfaces can reduce this burden by automating parts of the robot's behavior, increasing safety and comfort for everyone from operators conducting tight-tolerance assembly in manufacturing to home users of assistive robots.
Teleoperation is also being used for data collection of human demonstrations, both with simulated 
\cite{ROBOTURK-mandlekar18a,mees2022calvin,shridhar2022cliport,li2023behavior} and real robots \cite{RT1_brohan2022rt,RT2_rt22023arxiv,zhang2018deep}, 
to build datasets for use with imitation learning 
\cite{ROBOTURK-mandlekar18a,wang2023mimicplay} 
and offline reinforcement learning \cite{rosete2023latent,zhou2023real}.
Improvements to interfaces are required in order to make online teleoperation faster and more intuitive, as well as to improve the quality of trajectories for robot learning \cite{mandlekar2021matters,belkhale2023data}. 
Grasping and placing objects precisely and smoothly is still difficult for operators due to perception and haptic gaps \cite{zhu2023shared,qin2023anyteleop}.
Grasps often fail when small clearances aren't respected, and the limited visual cues afforded to operators can cause them to press objects down further than necessary when placing. 
 
\begin{figure}[t]
    \centering
    \includegraphics[width=.9\linewidth]{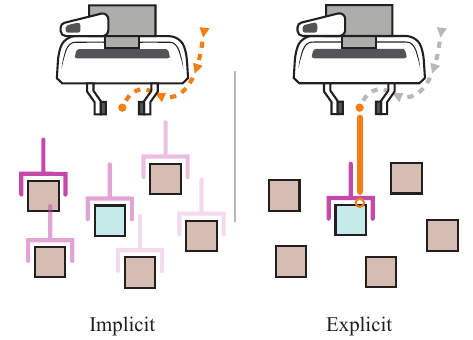}
    \caption{Implicit assistance (left) funnels the operator toward the goal predicted based on (for instance) the recent trajectory. The operator is not intended to change their input to influence the assistance. 
     Explicit assistance (right) affords the operator direct control over the inferred goal by pointing the gripper toward the object of interest. A local optimization selects a feasible, collision free pose.}
    \label{fig:imp_vs_exp}
\end{figure}

The foundation of most assistive teleoperation systems is prediction~\cite{dragan08,Hindsight,nikolaidis2017human}. 
Inferring, for instance, the operator's desired trajectory or end-effector goal based on their recent trajectory and context (i.e. scene, object, task) enables the automation of subsequent actions. 
Performant prediction systems can engage assistance fluently in proportion to their own confidence. The user teleoperates as they would without assistance. Their control over the predictions is \textit{implicit}, arising from how their state or actions correspond with some model of possible intended behavior. 

But the benefits of implicit inference-based assistance are difficult to realize in practice. 
Human environments pose challenges for online trajectory and goal prediction~\cite{lasota2017multiple,perez2015fast}. In clutter, where there are numerous possible target objects in close physical proximity, it is inherently difficult to predict manipulation targets as many goals may be consistent with a user's state or historical input.    
Poor predictions can lead the operator to modify their behavior in an attempt to better signal their goal---a confusing interaction, as the operator's mental model of the predictor is likely incorrect.
In such situations, it is preferable to provide an \textit{explicit} interface that accommodates the user's desire to exert direct control over the predicted intent.
Explicit input interfaces usually involve modal goal-specification interactions
which aren't suitable for online interaction, or additional input modalities, like natural language, that introduce complexity and potentially increase burden.

Our proposed interface for pick and place manipulation, shown in \figref{fig:imp_vs_exp}, leverages ``pointing'' of the end effector as an explicit input method, requiring neither an additional input modality nor a modal interaction. The approach offers assistance for a possible grasp or placement pose via optimization in a small region around where a ray from the gripper to the target object (grasping) or from the object in the gripper (placing) meets the scene geometry. Parallel computation allows the system to rank and filter many possible collision-free candidates and present suggestions that are responsive to the user's input at high frequency.

We implemented our proposed explicit interface on a simulated Franka Emika Panda robot and conducted a user study comparing it to an implicit-input assistive teleoperation method on pick-and-place stacking tasks with clutter. 
We find that operators prefer the explicit interface, experience fewer pick failures and report lower cognitive workload.
Our implementation of explicit assistance and study conditions in NVIDIA Omniverse Isaac Sim is available at \href{https://github.com/NVlabs/fast-explicit-teleop}{\texttt{github.com/NVlabs/fast-explicit-teleop}}.

\section{Related Work}

The design space for assistive teleoperation spans various types of operator input, different forms of assistance and a spectrum of manual to automatic engagement~\cite{dragan08}.

Most assistive teleoperation methods use a form of implicit input to autonomously generate improved robot actions. Early methods maintained a probability distribution over possible goals given users' recent actions and overrode user control with actions more in line with optimal trajectories to the inferred goal~\cite{dragan08,Hindsight,nikolaidis2017human}. 
When available, data enables the use of sophisticated predictive models like trajectory forecasting transformers~\cite{clever2021assistive} or multi-modal diffusion policies~\cite{yoneda2023diffusha}.
When a task reward is available, it is possible to use human-in-the-loop deep reinforcement learning~\cite{reddy2018shared}.
While some of these methods produce assistance based only on the current state, user interactions with the assistance are characteristically implicit as the user is not intended to control the state with the aim of modifying the assistance.

Human-in-the-loop autonomous systems commonly allow operators to explicitly specify goals, preview generated trajectories and supervise execution~\cite{leeper2012strategies,PerezDArpino2020ExperimentalAO}.
Most frequently, these interfaces use keyboard and mouse control over 6DOF interactive pose markers, enabling precise goal specification at the expense of fluency, making them unsuitable online continuous teleoperation. 

Assistance can also come in the form of augmented control input schemes.
\cite{losey22latent} used demonstrations to learn a task-specific low-dimensional control mapping, enabling operators to control a robot arm using only a 2D joystick.
\cite{cui23no} showed that such task-specific mappings can also be generated conditionally based on a language description of a task in a way that also allows natural language corrections during execution. 

Another approach is to dynamically constrain actions to, for example, avoid collisions with obstacles~\cite{rubagotti2019semi}, or reject probable inadvertent input in a fine manipulation task.
\cite{rosenberg93} introduced the concept of ``virtual fixtures,'' registered geometric overlays, typically specified beforehand using task knowledge, which produce sensory cues or alter control behavior as operators move through them.
These fixtures restrict motion within a region, like a virtual ruler confining end-effector motion to a line.

\section{Fast Explicit-Input Assistance}

We are interested in generating actions to assist a teleoperator.
Abstractly, the generation of these \textit{actions}---which may be poses, configurations or trajectories---is the result of an optimization based on \textit{state} information and \textit{context}:

\begin{equation}
    \text{actions} = \argmax_{\text{option} \in \mathcal{A}} f(\text{option}, \text{state}, \text{context})
    \label{assistance-formulation}
\end{equation}

The defining decisions we make about the implementation of \equref{assistance-formulation} that result in an effective explicit-input interaction are:
\begin{itemize}
    \item to use transparent and controllable state information;
    \item to prioritize smoothness of the assistance with respect to state in the selection of $f$. 
     Both the average and maximum variations in assistance for small state changes affect usability, as abrupt changes can be disorienting;
    \item and to treat the resulting action as a suggestion subject to user review and refinement. 
\end{itemize}

Conventional inference-based assistance systems attempt to represent the space of possible goal poses or next-actions in $\mathcal{A}$. They select for $f$ a model of the likelihood of the goal conditioned on the pose or recent trajectory of the robot. 

We similarly choose to produce useful poses for the operator, but we disregard the opaque history of the operator's actions and instead rely on immediately controllable present-state information. We leverage an intuitive ``pointing'' metaphor to allow the user to specify the anchor for a local optimization of an assistance pose. We define the optimization to be amenable to parallelization, ensuring it can compute at interactive speeds. The result is a pose suggestion that the user can ignore or modify by pointing the gripper before affirmatively accepting.

\subsection{Pointing as Ray Control}

\begin{figure}[t]
    \centering
        \includegraphics[width=.8\linewidth]{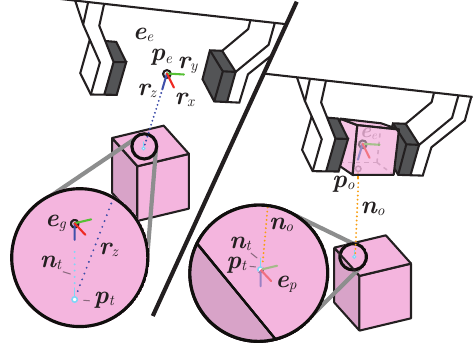}
    \caption{Our realizations of explicit grasping (left) and placing assistance (right) both center on the interaction of a ray from the gripper with scene geometry. A projected anchor pose is calculated then used to select amongst a set of candidate assistance poses.}
    \label{fig:assistance}
\end{figure}

Our experience is that the most understandable and controllable aspect of state is where the end-effector is pointing. 
Although pointing is governed by all six degrees of freedom (DoF) of the end effector pose---which we denote as $\boldsymbol{e}_e \in \textrm{SE}(3)$ located between the fingers---it particularly emphasizes control of the axis component $\boldsymbol{v} \in \mathbb{R}^3$ of the axis-angle $(\boldsymbol{v}, \theta)$ representation of the $\textrm{SO}(3)$ orientation. For convenience, we will also leverage the $\mathbb{R}^{4\times4}$ transformation matrix representation of the pose $\boldsymbol{e}_e$ consisting of rotation matrix $\boldsymbol{R} = [\boldsymbol{r}_x,\boldsymbol{r}_y,\boldsymbol{r}_z] \in \mathbb{R}^{3 \times 3}$ and translation component $\boldsymbol{p}_e \in \mathbb{R}^{3}$. We assign the $\boldsymbol{r}_z$ component outwards from the gripper, $\boldsymbol{r}_y$ perpendicular and along the axis of closing, and $\boldsymbol{r}_x$ perpendicular and away from the gripper camera. These axes are labeled on \figref{fig:assistance}.

Pointing the axis $\boldsymbol{r}_z$ is familiar for operators not only because it is a necessary component of most manipulation tasks, but also because it is a means to change the view of the ``eye-in-hand'' camera that is often available.
When unobstructed, this view is an innate visualization of the pointing input upon which crosshairs or a rendered lines can directly show the pointing axis.

The pointing target is the point $\boldsymbol{p}_t \in \mathbb{R}^3$ at which the ray extending from the end effector position $\boldsymbol{p}_e$ along $\boldsymbol{r}_z$ contacts the scene geometry, and $\boldsymbol{n}_t$ is the surface normal at the target point. These quantities can be approximated using depth data or a geometric representation that the robot has access to. They are simple to visualize by (for example) highlighting the point in a 2D view and drawing a tick mark in the normal direction. 

Previous works have focused on the proximity of the end effector to assistance candidates, but relative distinctions in distance are difficult to assess based on a remote 2D view. Proximity is chiefly a function of the 3DoF end effector position $\boldsymbol{p}_e$, whereas the axis $\boldsymbol{v}$ of the gripper orientation is characterized by just the 2 spherical coordinates, azimuth and elevation. Pointing does still usefully encode a nearness bias, since the area at which one can point at a given object is inversely proportional to the square of the distance to the object.
In other words, it is easy to point at things nearby, grows more difficult for things further away, and quickly becomes effectively impossible beyond a point. This bias is reinforced by the fact that one can only point at what can be seen and further objects are subject to greater occlusion.

\subsection{Grasp Pointing}

In order to suggest a possible grasp pose to the operator using our pointing interface we must define a mapping from the 6D pointing control to a 6D grasp pose. We denote the final grasp assistance suggestion as $\boldsymbol{e}^*_g$. 

A direct mapping would be to simply displace the current gripper orientation along the ray to some fixed offset of the target point, making no modification to the gripper orientation. 
However, our experience is that users often point at oblique angles but nonetheless desire an approach orthogonal to the object surface.
Instead of using $\boldsymbol{r}_z$, we use the negation of the surface normal $\boldsymbol{n}_t$ at the target point $\boldsymbol{p}_t$.

Users generally expect the angle $\theta$ about the axis to be the one that is ``most similar'' to their current orientation. To encode this geometrically, we project a reference vector anchored to the gripper onto the plane defined by the intersection point $\boldsymbol{p}_t$ and the normal vector $\boldsymbol{n}_t$. Any reference vector may be selected, however it is preferable to use one that is unlikely to be perpendicular to the plane, like $\boldsymbol{r}_x$ or $\boldsymbol{r}_y$. The minimal rotation is the geodesic between the current and the projected reference vector.

The resulting grasp anchor pose $\boldsymbol{e}_g$ provides an intuitive, cursor-like interaction when the gripper ray is swept across the scene. It is unlikely to be a satisfactory grasp on its own, however, because an orthogonal approach may be inappropriate for the object, or the position may cause contact with the object or other scene geometry. A generative grasp model can be used to provide a set $\mathcal{A}_{(\boldsymbol{e}_g)}$ of candidates near the anchor. The specification of ``near'' governs the smoothness of the assistance interaction, with smaller thresholds ensuring that the resulting poses do not change substantially as the cursor moves but necessarily excluding more suitable grasps that are too far away. Each candidate can be computed and scored independently, making this step highly parallelizable. The result nearest the anchor should be taken as grasping suggestion $\boldsymbol{e}^*_g$. Generally the quality and smoothness of the assistance improve as more candidates are considered so long as the computation runs at interactive rates.

\subsection{Placement Pointing}

As with grasp pointing, we seek a mapping from the 6D pointing control to a 6D end-effector placement pose, $\boldsymbol{e}^*_p$.

The object may have been grasped in an arbitrary orientation, so a direct mapping that translates the current pose along the gripper axis $\boldsymbol{r}_z$ toward the target point is unlikely to be useful for stably placing the object. Instead, we observe that the object was likely picked from a stable pose where it rested on a support facet defined by some point $\boldsymbol{p}_o$ and normal $\boldsymbol{n}_o$ pointing in the gravity direction. At the moment of the pick, the orientation of normal $\boldsymbol{n}_o$ can be recorded with respect to the end effector pose $\boldsymbol{e}_e$, and a point $\boldsymbol{p}_o$ on the object facet can be estimated by projecting the end-effector position $\boldsymbol{p}_e$ at the moment of the pick onto the scene geometry revealed after the object is lifted. 

It is now intuitive to map the control of the resulting plane $(\boldsymbol{p}_o, \boldsymbol{n}_o)$; the user principally controls the axis $\boldsymbol{n}_o$ to select a pose constrained to place the facet point $\boldsymbol{p}_o$ at the target point $\boldsymbol{p}_t$ and to align the object normal $\boldsymbol{n}_o$ opposite the target normal $\boldsymbol{n}_t$. Similar to the grasp mapping, the undetermined rotation of the object about the target normal is specified by finding the geodesic from a reference vector on the end effector (like $\boldsymbol{r}_x$ or $\boldsymbol{r}_y$) to the same vector projected onto the target plane.

The resulting placement anchor pose $\boldsymbol{e}_p$ is a direct, cursor-like projection of the grasped object into a placement, and is used in a similar manner as the grasp anchor pose. The anchor itself may not be a feasible placement pose if it puts the object or the gripper into contact with the scene.
Candidates $\mathcal{A}_{(\boldsymbol{e}_p)}$ can be generated in the local region around the anchor using any generative object placement method, with the candidate nearest the placement anchor pose serving as the suggestion $\boldsymbol{e}^*_p$ to the user. 

\begin{figure*}%
\captionsetup[subfigure]{labelformat=empty}
    \centering
    \begin{minipage}{0.49\linewidth}
    \subfloat[\label{fig:camera-views}]{%
    \includegraphics[width=\linewidth]{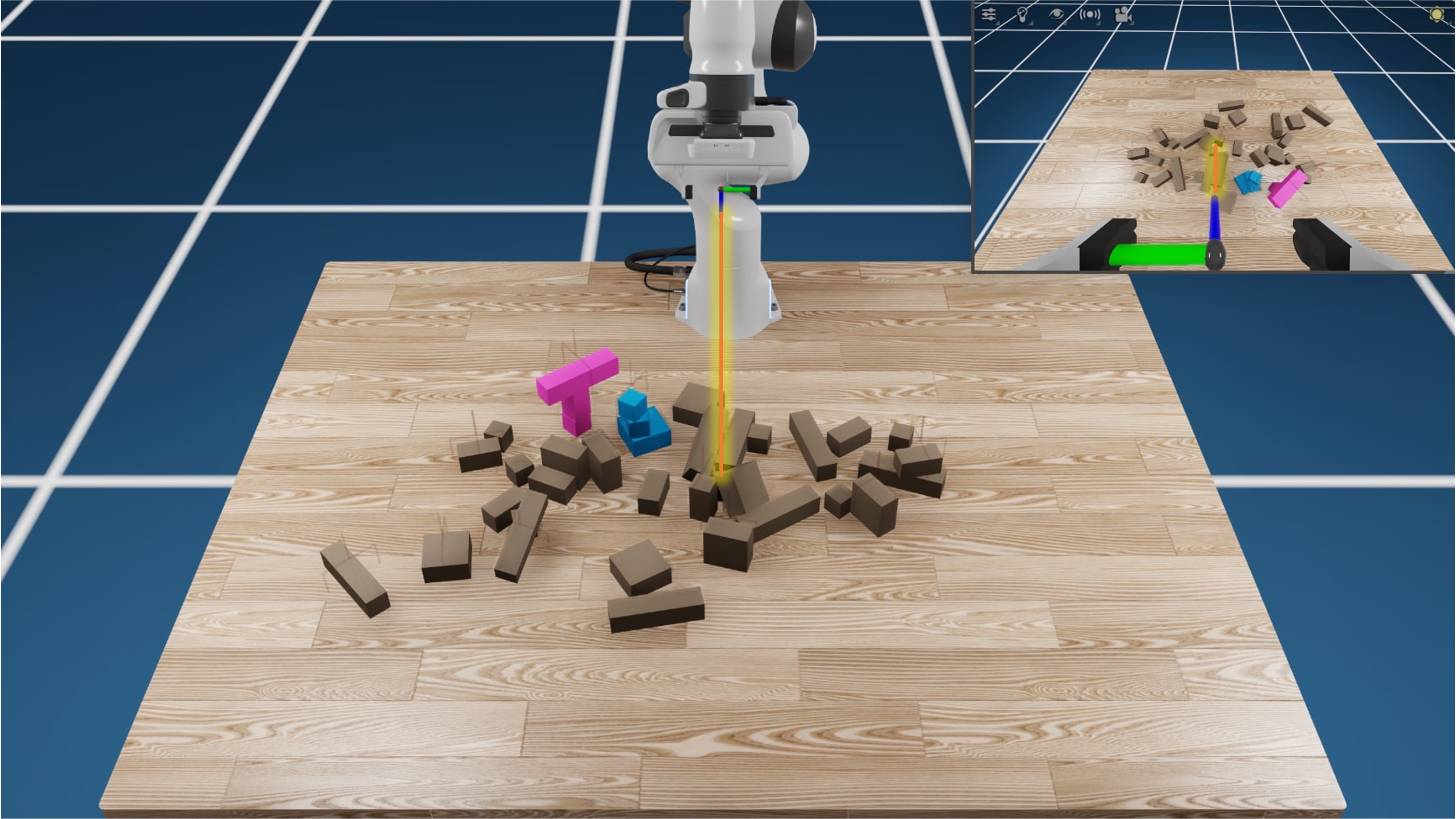}
    }%
    \end{minipage}
    \hspace{10pt}
    \begin{minipage}{.41\linewidth}
    \vspace{-9pt}
            \subfloat[\label{fig:viz-grasp}]{\includegraphics[width=.45\linewidth]{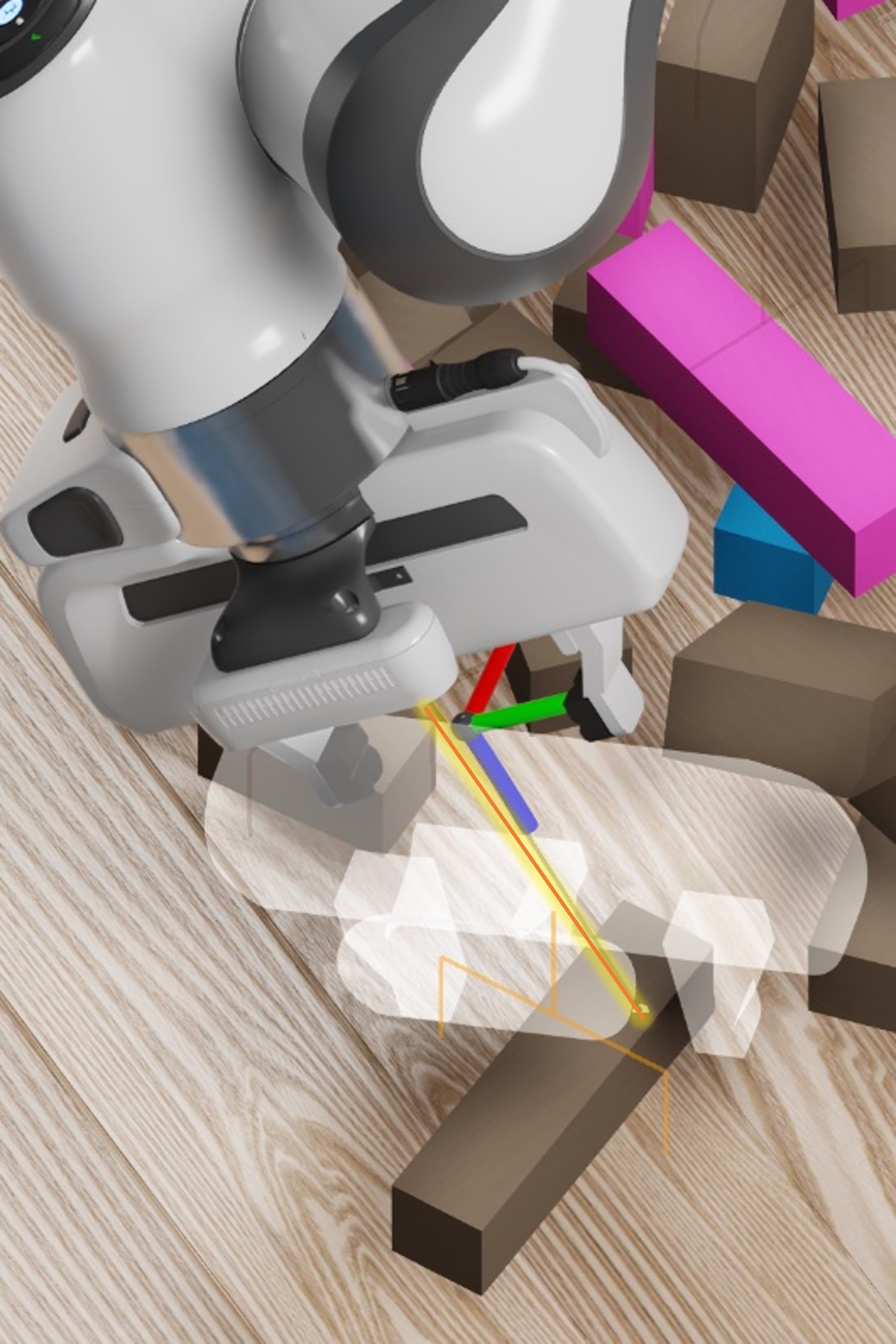}}\hspace{5pt}
    \subfloat[\label{fig:viz-place}]{\includegraphics[width=.45\linewidth]{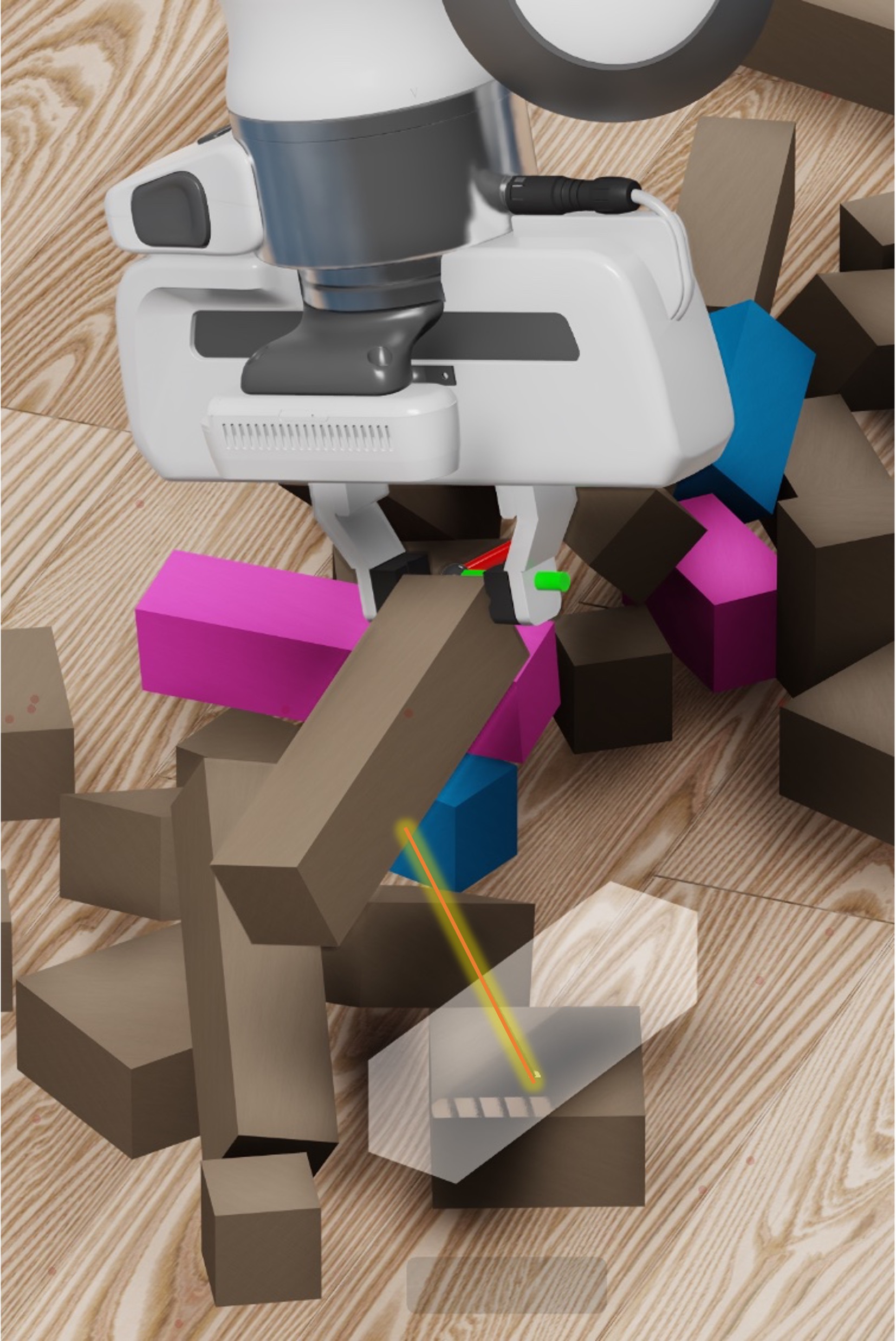}}
    \end{minipage}
    \begin{minipage}{1\linewidth}
    \centering
    \vspace{-20pt}
    \subfloat[\label{fig:task-environments}]{\includegraphics[width=.3\linewidth]{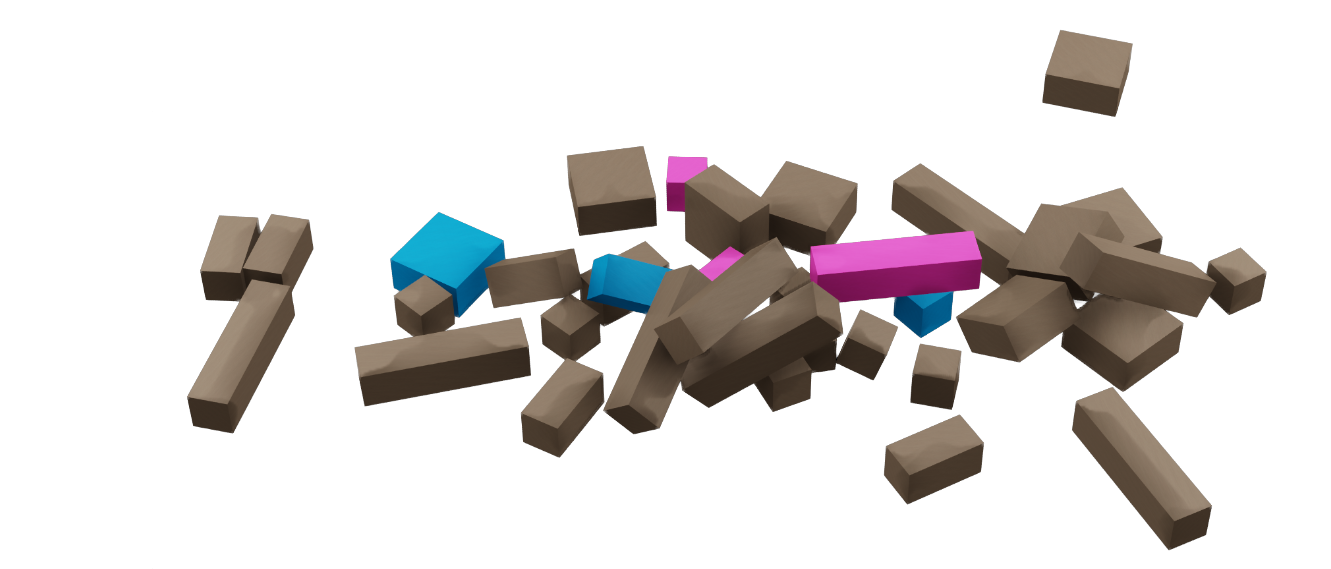}}
    \subfloat[]{\includegraphics[width=.3\linewidth]{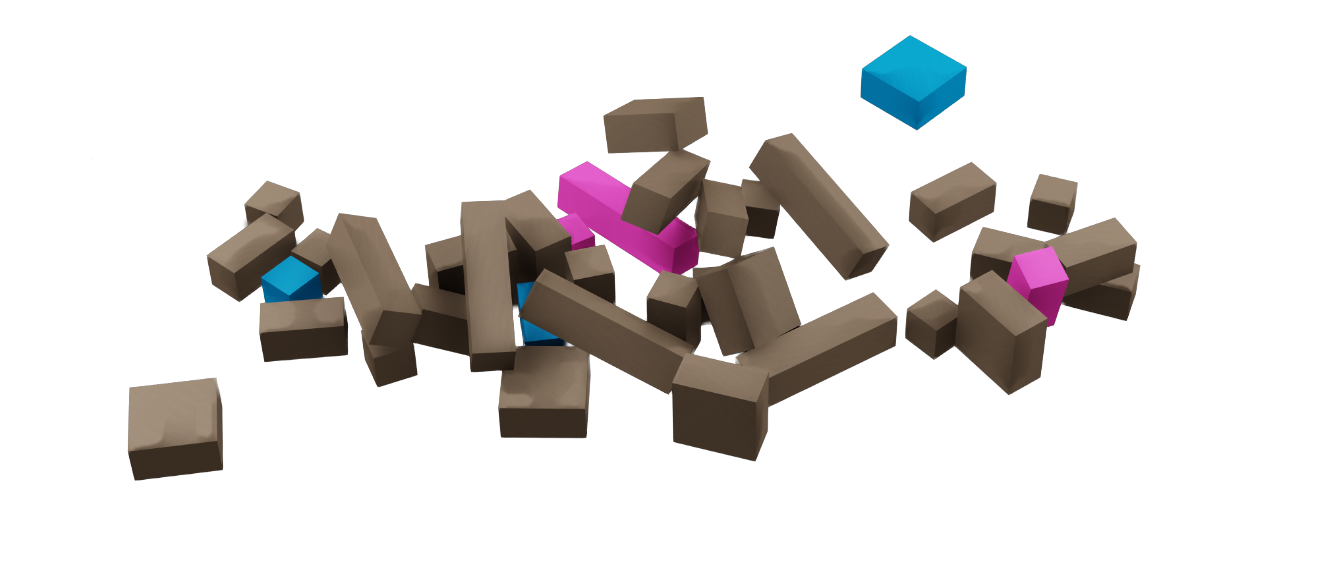}}
    \subfloat[]{\includegraphics[width=.3\linewidth]{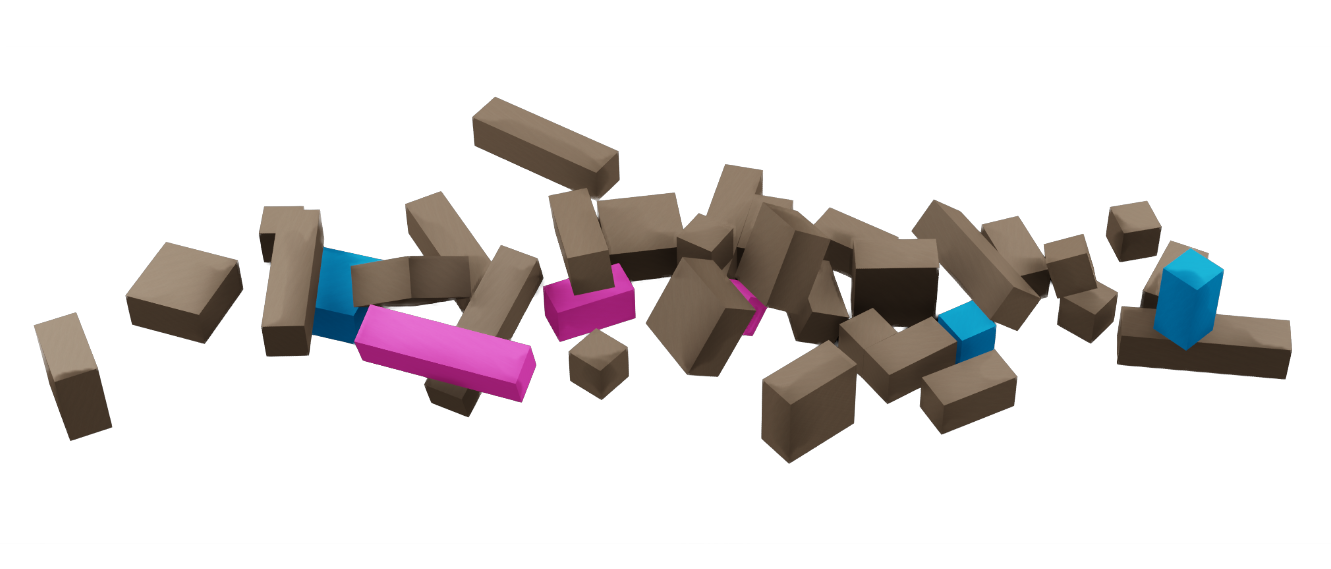}}%
    \end{minipage}
    
    \centering

    \caption{ 
    The operator controls the robot while looking at two camera views displayed picture-in-picture (left). Assistance suggestions are shown as a ``ghost gripper'' for grasping and a ``ghost shape'' for placing actions (right). Ray visualizations are exaggerated for legibility in print. The experimental task involved participants extracting and stacking blue and pink blocks that were initially scattered in one of three clutter configurations (bottom).}
   \label{fig:system}

\end{figure*}

\subsection{Snapping}\label{sec:snapping}

As a consequence of prioritizing responsiveness, the range of inputs which our methods map to any particular assistance anchor pose $\boldsymbol{e}_g$ or $\boldsymbol{e}_p$ is small. 
Certain ``easy'' poses like a perfectly aligned side-grasp might be frustratingly difficult to specify. We use \textit{snapping} to nudge the generated assistance toward these preferred poses, providing the flexibility to control the grasp suggestion (as is typically needed in cluttered scenes) or to easily snap into commonly used grasps when feasible. The behavior of snapping is demonstrated in the accompanying video. 

Snaps are encoded by one or more potential fields $\phi(\cdot)$ over poses. 
After anchor poses $\boldsymbol{e}_g$ or $\boldsymbol{e}_p$ are calculated, a local optimization over $\phi$ occurs, checking to see if there is a lower potential pose within an $\epsilon$ distance threshold that would breach potential threshold $\gamma$.
If so, candidates from $\mathcal{A}_{(\boldsymbol{e}_g)}$ or $\mathcal{A}_{(\boldsymbol{e}_p)}$ are ignored and the snap pose is provided as the suggestion.

In practice, we find that specifying a set of poses that align with object centroids coupled with proximity potential $\phi(e^*) = \min_{G_i \in G}d(\boldsymbol{e}^*, G_i)$ is useful for picking and placing and requires no additional task context.

Following \cite{mazzotti2016measure}, we define the distance between the poses $\boldsymbol{x}, \boldsymbol{y} \in \textrm{SE}(3)$ with position components $\boldsymbol{p}_x, \boldsymbol{p}_y \in \mathbb{R}^3$ and rotational components $\boldsymbol{R}_x, \boldsymbol{R}_y \in \mathbb{R}^{3\times3}$, as
\begin{equation}
        d(\boldsymbol{x},\boldsymbol{y})^2 = ||\boldsymbol{p}_x - \boldsymbol{p}_y||_2 + 2\beta^2 (1 - \frac{tr(\boldsymbol{R}_y^{-1}\boldsymbol{R}_x)}{3})),
\end{equation} where $\beta$ weights the translation and rotation contributions to the distance.

\section{Experiment}

We conducted a within-subjects user study where participants completed stacking tasks without assistance (\textbf{CON}), with implicit inference-based assistance suggestions (\textbf{IMP}), and with explicit-input assistance suggestions (\textbf{EXP}).

Participants completed a multi-step singulation and stacking task where they created multiple stacks of particularly-colored blocks from a cluttered pile. The task was designed to have few prescribed steps and many possible intermediate goals.

We expected that participants would:
\begin{itemize}
    \item[\textbf{H1}]: be most effective at completing the task using EXP.
    \item[\textbf{H2}]: make most use of suggestions provided by EXP
    \item[\textbf{H3}]: report the lowest workload when using EXP.
    \item[\textbf{H4}]: feel that the suggestions from EXP better match their preferences.  
    \item[\textbf{H5}]: feel that they understand the behavior EXP better than that of IMP.
\end{itemize}

\subsection{System}

Participants interact with a Franka Panda robot simulated in NVIDIA Omniverse Isaac Sim. 
Grasp sampling and collision checking operations are GPU accelerated using NVIDIA Warp~\cite{warp2022}.

\paragraph{Input} Users provide input using a 6DOF mouse, a spring-suspended puck that they can displace in three spatial dimensions while simultaneously panning, tilting, or twisting to provide 3D rotation~\cite{dhat2024using}. 

\paragraph{Robot Control} User input is interpreted as a twist goal for the robot's end-effector. We integrate the twist over a fixed timestep, apply the resulting transformation to the current end-effector pose, and provide the result as a pose goal to the robot controller, a Riemannian Motion Policy implemented in RMPFlow~\cite{rmpflow}. To avoid large accelerations, the pose goal is passed through a low pass filter.

\paragraph{Camera Control} Users operate the robot while monitoring a fixed view, showing most of the robot and the workspace, and a dynamic view affixed to the gripper pointing towards the fingers. As shown in~\ref{fig:system}, one view is foregrounded at a time, and user input is interpreted in the frame of the foregrounded view.

\paragraph{Assistance} Offers of assistance are visualized as ``ghosts,'' as shown in \figref{fig:system}. Holding a button on the 3D mouse engages the assistance, forwarding the suggested pose as a goal for the controller with an additional preprocessing step to ensure poses are approached from the front.

\begin{figure}[!t]
    \centering
    \includegraphics[width=\linewidth]{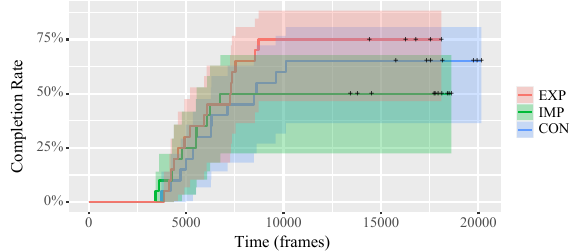}
    \caption{Survival analysis ($\uparrow$) of participant's completion of the task over time. Lines plot percentage of participants that completed the task at the time and Xs mark termination without completion. Differences lie within the 95\% confidence interval, with a trend that the probability of having completed the task grows most quickly for the explicit input interface and reaches a higher peak.}
    \label{fig:survival-analysis}
\end{figure}

\paragraph{Explicit Assistance Condition (EXP)}

We implement our grasp pointing assistance approach using a simple approach-vector parameterized sampling scheme, looking for the nearest non-colliding pose amongst 7125 translated and rotated candidates around the grasp anchor pose. The samples are distributed in a fixed 2cm diameter, 1cm thick disc pattern. We did not use a placement sampler as we assessed that direct control over the placement anchor pose was sufficient for the experimental task. Raycasting is performed against a mesh representation of the scene. We generate axis aligned grasp and placement poses and use them to define a snapping potential as described in \secref{sec:snapping}.

\paragraph{Implicit Inference-Based Assistance Condition (IMP)}
Following \cite{dragan08}, we attempt to infer the user's goal by selecting the most-probable goal $G^*$ from a predefined set of candidates $\mathcal{G}$ based on a recent window of the robot's trajectory $\xi_{S\rightarrow U}$ from start pose $S$ to current pose $U$:

\begin{equation}
    G^* = \argmax_{G \in \mathcal{G}} \bigg(\frac{e^{-C_G({\xi_{S\rightarrow U}}) - C_G(\xi^*_{U\rightarrow G})}}{e^{-C_G(\xi^*_{S\rightarrow G})}} \cdot e^{-d(U,G)} \bigg)
\end{equation}

The first term assigns greater likelihood to goals for which the user's trajectory, completed optimally by $\xi^*_{U\rightarrow G}$, has cost similar to the cost of the optimal trajectory $\xi^*_{S\rightarrow G}$. The second term serves as a prior, assigning more mass to goals that are closer to current pose.
We use $C_G(\xi_{X\rightarrow Y}) =  d(X,Y)^2$ and reset $S$ if 2 seconds pass with no control input. The same set of axis-aligned grasp and placement poses used for snaps are used as $\mathcal{G}$, and collision checking is performed across this set to ensure no in-collision poses are offered.

\begin{table}[!t]
\centering
\footnotesize
    \caption{Comparison of Condition Preference \underline{C}ounts $\uparrow$}
  \label{tab:preferences}
\begin{tabular}{c c c c c c c}
    \toprule
    & A    & B   & \mc{$C_A$} & \mc{$C_B$} & \mc{$\frac{C_A}{C_A+C_B}\%$}(CI)          & \mc{$p$} \\
    \midrule
    \colrot{3}{EQ1} & 
    EXP  & CON  & 11 & 1 & 92  (62, 100) & \bd{.010} \\
    & \ditto    & IMP  &  \ditto  & 8 & 60  (34, \phantom{0}80) & .648 \\
    & CON & \ditto  & 1  & \ditto & 11  (\phantom{0}0, \phantom{0}48) & .078 \\
    \midrule
    \colrot{3}{EQ2} & 
    EXP  & CON  & 10 & 3 & 79  (49, \phantom{0}95) & .277 \\
    &  \ditto   & IMP  &  \ditto  & 7 & 61  (33, \phantom{0}82) & .688 \\
    & CON & \ditto  & 3  & \ditto & 30  (\phantom{0}7, \phantom{0}65) & .688 \\
    \midrule
    \colrot{3}{EQ3} & 
    EXP  & CON  & 14 & 1 & 94  (68, 100) & \bd{.003} \\
    &  \ditto   & IMP  &  \ditto  & 5 & 75  (49, \phantom{0}91) & .127 \\
    & CON & \ditto  & 1  & \ditto & 17  (\phantom{0}0, \phantom{0}64) & .219 \\
    \midrule
    \colrot{3}{EQ4} & 
    EXP  & CON  & 14 & 2 & 88  (62, \phantom{0}98) & \bd{.013} \\
    &  \ditto   & IMP  &  \ditto  & 4 & 79  (52, \phantom{0}94) & .061 \\
    & CON & \ditto  & 1  & \ditto & 33  (\phantom{0}4, \phantom{0}78) & .687 \\
    \bottomrule
\end{tabular}

\end{table}

\subsection{Procedure}

Participants were told they would use a 3D mouse to control a robot with three different systems, some of which would provide suggestions they could use to help them complete tasks.
Each session began with an interactive 3D mouse tutorial, followed by a robot control tutorial where they had to grasp and lift a block, and finally an assistance tutorial which demonstrated what suggestions of assistance would look like and how to use them.

For each condition, participants were given a brief verbal introduction to how the system would behave and asked to ``warm up'' by stacking a block.
Once satisfied that they understood the system, participants completed a single stack task for 3 minutes, then had a maximum of 7 minutes to complete the multi-step stacking task.
A post-interaction survey included the NASA-TLX questionnaire~\cite{hart1988development}, three agreement questions regarding their sense of control over the suggestions (reported as assistance composite) and one regarding their sense of understanding. Rating questions were represented using 7--point scales.

A final set of forced-choice questions probed which system ``felt easiest to use" (EQ1), and which system had the suggestions that ``made it easiest to do the task the way [they] wanted to" (EQ2) which they best understood ``why [the suggestions] behaved the way they did" (EQ3), and ``felt most in control of" (EQ4). Finally, participants completed demographic questions and rated their familiarity with robots, operating robot arms, 3D mice, and playing video games.
Sessions lasted between 45-60 minutes total.

\paragraph{Participants} We recruited 20 participants (18 male, 2 female, aged 19-39 \textit{M}=25.1, \textit{SD}=5.45) from the University of Washington under an IRB approved study plan. Many participants were roboticists, rating their familiarity with robots highly (\textit{M}=4.80, \textit{SD}=2.08, 7--point scale). Only two participants reported being familiar with 3D-mice (rating $>$4 on 7--point scale). All participants were right handed.

\subsection{Methods}

We analyze logged events, survey data and supplemental annotations using generalized linear mixed models to account for inter- and intra-participant variance. The effect of an experimental condition is given as either a ratio or difference of the estimated marginal mean against another contrasting condition, and significance is determined using 95\% confidence intervals. We conducted survival analysis to characterize task completion rates over time. Statistical details are reported in the supplementary materials.

\subsection{Results}

\textbf{H1}: Participants experienced significantly fewer failed picks in EXP when compared to IMP or CON, and there was a trend indicating that they experienced fewer place failures as well, as shown in \tabref{tab:failures}. There were trends indicating that users of the explicit interface complete the task with higher frequency and stack objects more quickly, as shown in \figref{fig:survival-analysis}, however the differences are not statistically significant. 

\textbf{H2}: There was no measurable difference in the duration or number of engagements of the assistance between the implicit and explicit interfaces. Qualitatively, we observed that some participants made use of the explicit assistance system without engaging it. 

\textbf{H3}: Mean workload was lowest for the explicit condition, however the difference was only significant when compared to the control condition. The implicit input condition was rated as higher workload than the explicit condition and lower than no assistance at all, however these differences were not statistically significant, as shown in \tabref{tab:tlx}.

\begin{figure}
    \centering
    \hspace{-.3cm}
    \includegraphics{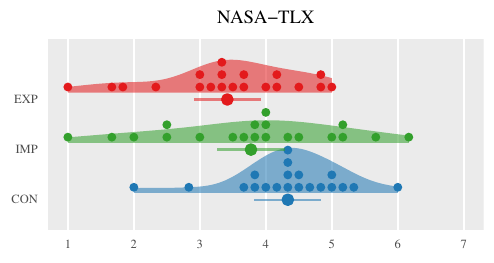}
    \centering
    \includegraphics{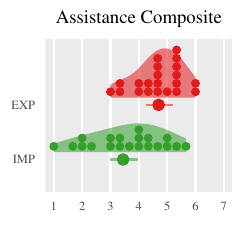}
    \includegraphics{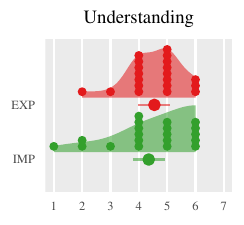}
    \vspace{-.1cm}
    \caption{Raw data for subjective scores collected on 7--point scale with density estimates overlaid. Point and bar show estimated marginal mean with 95\% confidence interval.}
    \label{fig:assistance-subjective}
\end{figure}

\textbf{H4}: Participants indicated that the explicit assistance interface was more controllable, rating it 1.25 points (CI .52, 1.99) more highly on average on our assistance composite scale (reported in \tabref{tab:assistance-subjective} and \figref{fig:assistance-subjective}). 

\textbf{H5} Participants rated their understanding higher on average, but the difference was not statistically significant as shown in \tabref{tab:assistance-subjective} and \figref{fig:assistance-subjective}.

\begin{table}[!t]
\centering
\footnotesize
    \caption{NASA-TLX scores $\downarrow$}
  \label{tab:tlx}
  \begin{tabular}{@{\hspace{4pt}}c @{\hspace{4pt}}c *{2}{@{}c} c c}
\toprule
A                 & B   & \mc{$M_A$($SE_A$)} & \mc{$M_B$($SE_B$)} & \mc{$M_A - M_B(CI)$} & \mc{$p$} \\
\midrule

EXP     & IMP  & 3.42 (.25) & 3.77 (.25) & -.36 (\phantom{0}-.97, \phantom{0}.25) & .335 \\
\ditto        & CON  &     \ditto        & 4.33 (.25) & -.92 (-1.53,\phantom{0}-.31) & \bd{.002} \\
IMP     & \ditto  & 3.77 (.25) & \ditto & -.56 (-1.17, \phantom{0}.05) & .079 \\

\bottomrule

\end{tabular}
\end{table}

\begin{table}[!t]
\footnotesize
    \caption{Assistance Subjective Scores $\uparrow$}
  \label{tab:assistance-subjective}
  \begin{tabular}{l @{\hspace{-4pt}}c @{}c @{}c @{}c @{}c @{}l}
\toprule
& \multicolumn{2}{c}{EXP} & \multicolumn{2}{c}{IMP} \\
Question   & \mc{$M_A$} & \mc{$SE_A$} & \mc{$M_B$} & \mc{$SE_B$} & \mc{$M_A - M_B(CI)$} & \mc{$p$} \\
\midrule
Composite     & 4.70 & .253 & 3.44 & .253 & 1.25 (\phantom{0-}.52, 1.99) & \bd{.002} \\
Understanding & 4.62 & .274 & 4.29 & .274 & \phantom{0}.33 (\phantom{0}-.47, 1.14) & .398 \\
\bottomrule

\end{tabular}
\end{table}

\begin{table}[!t]
\footnotesize
    \caption{Failure Counts $\downarrow$}
  \label{tab:failures}
  \begin{tabular}{@{\hspace{2pt}} l @{\hspace{4pt}} c @{\hspace{4pt}} c c c c c}
    \toprule
    & A                 & B   & \mc{$M_A$($SE_A$)} & \mc{$M_B$($SE_B$)} & \mc{$M_A/M_B(CI)$} & \mc{$p$} \\
    \midrule
     \colrot{3}{Pick} & EXP     & IMP  & 1.13 (\phantom{0}.32) & 2.48 (\phantom{0}.59) & \phantom{0}.46 (\phantom{0}.23, \phantom{0}.91) & \bd{.028} \\
    &   \ditto     & CON &      \ditto        & 4.22 (1.15) & \phantom{0}.27 (\phantom{0}.10, \phantom{0}.75) & \bd{.008} \\
    &IMP      & \ditto & 2.48 (\phantom{0}.59)  & \ditto & \phantom{0}.59 (\phantom{0}.27, 1.27) & .242 \\
    \midrule
    
    \colrot{3}{Place} & EXP     & IMP  & \phantom{0}.55 (\phantom{0}.18) & \phantom{0}.59 (\phantom{0}.18) & \phantom{0}.93 (\phantom{0}.38, 2.29) & .980 \\
     &   \ditto    & CON &     \ditto         & 1.22 (\phantom{0}.30) & \phantom{0}.45 (\phantom{0}.20, \phantom{0}.98) & \bd{.043} \\
     & IMP      &  \ditto & \phantom{0}.59 (\phantom{0}.18)  & \ditto & \phantom{0}.48 (\phantom{0}.23, 1.04) & .065 \\

  \bottomrule
\end{tabular}
\end{table}

\section{Discussion and Limitations}

We designed an explicit-input teleoperation interface that is interpretable, responsive, unobtrusive and capable. These design principles have inherent tradeoffs. For example, making assistance more capable may result in a less responsive and less usable system. Reducing latency is only desirable if interpretability can be maintained, a trade-off that often appears when considering how to configure anytime sampling-based planners.

Our implementation is deployed in simulation, making it applicable to simulated data collection or robot teaching interactions. Porting our system to teleoperation of a real robot would require the integration of appropriate generative grasp- and placement-pose models, as well as object state estimation or point cloud-based occupancy checking methods. Our experimental assessment of the interface informs and motivates the future development of physical implementations. Future work should also explore placement assistance with objects and support surfaces that are not well-approximated as planes.

\section{Conclusion} 
\label{sec:conclusion}

We contribute a new framing for assistance interactions based on explicit input and two new algorithms and interfaces for online teleoperation, designed to leverage GPU-based parallel computation to calculate grasping and placing feasible options online---even in clutter.
Our work goes beyond individual picks by also considering assistance during placement, thus offering a complete workflow for multi-step pick and place tasks. 
The results of our study highlight the promise of this new kind of assistance interaction, and motivate us to further explore how accelerated computation can augment teleoperation.

\section*{Appendix A}
\section*{System}
\addcontentsline{toc}{section}{Appendix A: System}
All participants interacted with the simulation running in NVIDIA Omniverse Isaac Sim 2022.2.0 on a machine with an RTX 3060 Ti. The simulation ran at interactive framerates, averaging about 38fps and dipping to lows of about 15fps when participants created large numbers of contacts by pushing through many blocks at once. 

\begin{figure}[t]
    \centering
    \includegraphics[width=0.9\linewidth]{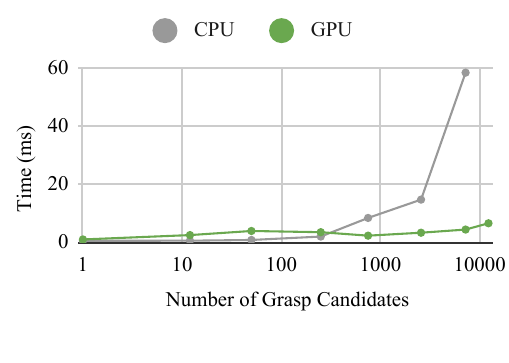}
    \vspace{-0.7cm}
    \caption{GPU acceleration is necessary to check thousands of candidate grasp poses for scene collisions while maintaing system responsiveness. GPU results from an NVIDIA RTX 3060 Ti, compared to single-threaded CPU execution on an AMD Ryzen 9 5900X.}
    \label{fig:performancegraspsampler}
\end{figure}

The bottleneck computation in the assistance systems was the filtering step where generated poses were rejected if they created collisions between the gripper and the scene. GPU acceleration was necessary for considering thousands of candidates each frame, as shown in the performance comparison in \figref{fig:performancegraspsampler}. Checking that candidates had feasible inverse kinematics solutions was not feasible at the time of the study. Participants encountered unreachable suggestions only infrequently because the block scatterings were placed comfortably within the robot's workspace. Participants that knocked or placed blocks further away were more likely to encounter unreachable suggestions.

\begin{figure}[t]
    \centering
    \includegraphics[width=0.8\linewidth]{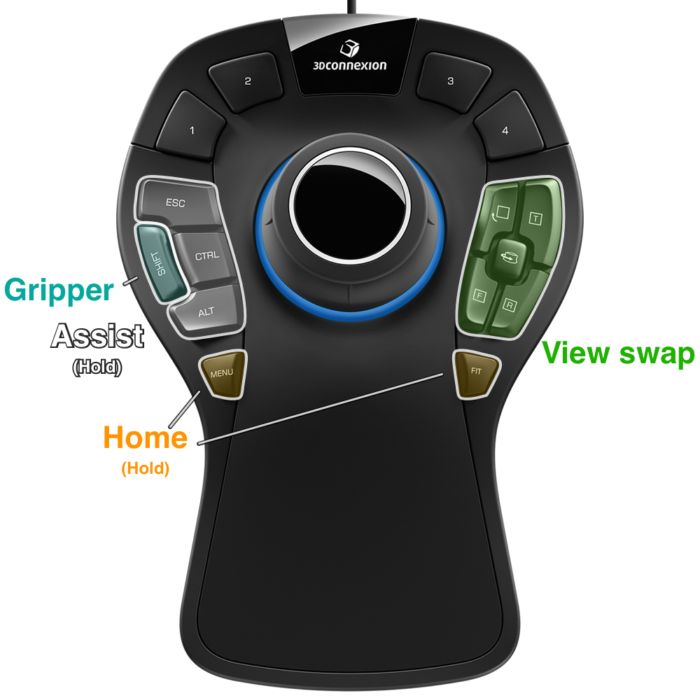}
    \caption{The mapping of buttons to system controls used during the study.}
    \label{fig:input-mapping}
\end{figure}

A 3DConnexion SpaceMouse Pro was used for all control input. The input mapping used was provided to participants as a printout (shown in \figref{fig:input-mapping} for reference during the study.

\section*{Appendix B}
\section*{Trajectory Labeling}
\addcontentsline{toc}{section}{Appendix B: Trajectory Labeling}

\begin{table}[t]
\centering
\caption{Event Codes and Descriptions}
\label{tab:annotations}
\begin{tabularx}{\linewidth}{>{\ttfamily}l X}
\toprule
\textbf{Code} & \textbf{Description} \\
\midrule

\multicolumn{2}{l}{\textbf{Tasks}} \\ 
pickt    & Successful pick: task block. \\ 
picko    & Successful pick: other block. \\
release  & Release with support. Code moment of release, then code outcome when it is clear. \\ 
drop     & Drop (the block in the gripper). Code moment of release, then code outcome when it is clear. \\ 
place    & Successful place: Stable place (of object in gripper). Code moment object stable at rest. \\ 
\addlinespace
\multicolumn{2}{l}{\textbf{Errors}} \\
erkno    & Deconstructed (``knocked over") existing tower. \\ 
erbut    & Unintentional drop, likely due to accidental button press. \\ 
erair    & Unsuccessful pick attempt, air grasp. \\
erpop    & Unsuccessful pick attempt, pop or slip out of gripper. \\ 
erfal    & Unsuccessful place attempt, contacted tower but fell. \\
erpmi    & Unsuccessful place attempt, missed tower. \\
ercon    & Object lost from gripper due to contact with scene. \\ 
\addlinespace
\multicolumn{2}{l}{\textbf{Milestones}} \\ 
blue1    & Assumed at start. Code after errors that cause deconstruction. \\
blue2    &  \\
blue3    &  \\
pink1    & Assumed at start. Code after errors that cause deconstruction. \\
pink2    &  \\
pink3    &  \\
\bottomrule
\end{tabularx}
\end{table}

Participant stacking trajectories were manually annotated by two of the authors using ELAN~\cite{ELAN2024}. The events annotated and their descriptions are given in \tabref{tab:annotations}. Only a subset of the events were analyzed for this work.

\section*{Appendix C}
\section*{Statistical Details}
\addcontentsline{toc}{section}{Appendix C: Statistical Details}

All analyses were conducted in R. Linear mixed models (LMMs) and generalized linear mixed models (GLMMs) were fit using \texttt{lme4}~\cite{Bates2015} or \texttt{glmmTMB}~\cite{brooks2017glmmtmb} when modeling zero-inflated distributions. Estimated marginal means were computed using \texttt{emmeans}~\cite{lenth2024}. Exploratory factor analysis was conducted using \texttt{psych}~\cite{revelle2024}, and survival analysis was conducted using \texttt{survival}~\cite{therneau2024survival}.

Where presented, models are described in Wilkinson notation~\cite{Wilkinson1973}. Linear mixed models were used for continous response variables, and Poisson GLMMs with a log link function were used to model count data. For linear mixed models, the Kenward-Roger method of estimating degrees of freedom was used.

\subsection{Subjective Assistance Scores}

Users responded to four Likert items after experiencing assisted conditions (either IMP and EXP). The items' statements are shown in \tabref{tab:questions}. We conducted an exploratory factor analysis (EFA) using the minimum residual method to identify the structure of responses to these novel items. The EFA indicated that a one-factor solution was sufficient, however item Q9 showed low communality, so it was analyzed as a standalone item, ``understanding.'' Responses to the remaining items were averaged to form the ``assistance composite'' score. 

The responses for the composite scale and standalone item were both independently analyzed using a LMM that accounted for the condition as well as inter-participant differences.

\[
\text{SubjectiveScore} \sim  \text{Condition} + (1 | \text{Subject})
\]

The resulting estimated marginal means are shown in \tabref{tab:assistance-subjective}. Means were compared using \textit{t} tests.

\begin{table}[t]
\centering
\caption{Survey Questions and Codes}
\label{tab:questions}
\begin{tabularx}{\linewidth}{>{\raggedright\arraybackslash}p{1.5cm} X}
\toprule
\textbf{Code} & \textbf{Description} \\
\midrule

\multicolumn{2}{l}{\textbf{NASA TLX}} \\
Q1 & How mentally demanding was the task? \\
Q2 & How physically demanding was the task? \\
Q3 & How hurried or rushed was the pace of the task? \\
Q4 & How successful were you in accomplishing what you were asked to do? \\
Q5 & How hard did you have to work to accomplish your level of performance? \\
Q6 & How insecure, discouraged, irritated, stressed, and annoyed were you? \\
\addlinespace

\multicolumn{2}{l}{\textbf{Agreement}} \\
Q7  & “The suggestions made it easy to accomplish the task." \\
Q8  & “The suggestions made it easy to accomplish the task in the way that I wanted." \\
Q9  & “I understood why the suggestions behaved the way they did." \\
Q10 & “I was in control of the suggestions." \\
\addlinespace

\multicolumn{2}{l}{\textbf{Open-ended}} \\
Q11 & Briefly describe the strategy you used for completing the task. \\
Q12 & What were your biggest frustrations with this system? \\
\addlinespace

\multicolumn{2}{l}{\textbf{Concluding Questions}} \\
EQ0  & Which system was most effective for the task? \\
EQ1  & Which system felt easiest to use? \\
EQ2  & Which system's suggestions made it easiest to do the task the way you wanted to? \\
EQ3  & With which system did you best understand why the suggestions behaved the way they did? \\
EQ4  & With which system did you feel most in control of the suggestions? \\
EQ5 & What were the major reasons for your choices? \\

\bottomrule
\end{tabularx}
\end{table}

\subsection{Condition Preferences}

\begin{table}[!t]
\footnotesize
    \caption{Comparison of condition preference counts for EQ0}
    \label{tab:eq0}
\begin{tabular}{c c c c c c @{}c c}
    \toprule
    & A    & B   & \mc{$C_A$} & \mc{$C_B$} & \mc{$\frac{C_A}{C_A+C_B}\%$} & \mc{CI}          & \mc{$p$} \\
    \midrule
    \colrot{3}{EQ0} & 
    EXP & CON  & 12 & 1 & 92  & (64, 100) & \bd{.010} \\
    & \ditto & IMP  & \ditto & 7 & 63  & (38, \phantom{0}84) & .359 \\
    & CON & IMP  & 1  & \ditto & 13  & (\phantom{0}0, \phantom{0}53) & .141 \\
    \bottomrule
\end{tabular}
\end{table}

For each of the forced-choice preference questions, a multinomial test was performed to evaluate whether participants’ preferences among the three conditions were evenly distributed. Tests for questions EQ1, EQ3, and EQ4 were significant, while a test of EQ2 was not.

Pairwise binomial tests were conducted to compare preferences between each pair of conditions, using Holm-Bonferroni corrections to account for multiple comparisons. The results of the pairwise comparisons are shown in \tabref{tab:preferences}. Responses to question EQ0 were highly similar to that of the other questions, but are given in separate table \tabref{tab:eq0} for completeness.

\subsection{Pick Failure Count}

\begin{figure}
    \centering
    \includegraphics{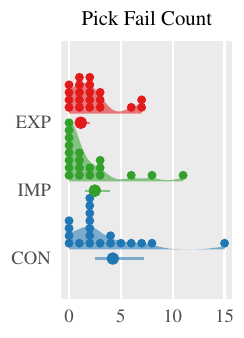}
    \includegraphics{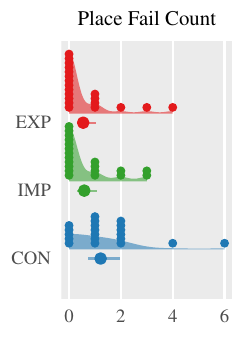}
    \caption{Counts of pick and place errors observed per participant by condition. Point and bar show estimated marginal mean and 95\% confidence interval.}
    \label{fig:failures}
\end{figure}

Pick failure models incorporated order, condition and block configuration factors as well as participant random effects.

A GLMM with a Poisson link function was used to model the count data. Excess zeros were observed in the baseline condition (CON), so a zero-inflation binomial term with the condition as the sole fixed effect was incorporated.

\begin{align*}
\text{PickFailureCount} \sim &\ \text{Order} * \text{Condition} \\
                             &+ \ \text{Configuration} + (1 | \text{Subject})
\end{align*}

A Type III Wald chi-square test was conducted to examine the effects of order, condition, environment, and their interaction on the number of pick failures. The main effect of order was significant, $\chi^2(2)=11.40,p=.003$, indicating that the number of pick failures differed depending on the order the condition was experienced in, consistent with a learning effect. The main effect of condition was also significant, $\chi^2(2)=14.10,p<.001$, suggesting differences in pick failures across conditions. The main effect of environment was not statistically significant, $\chi^2(2)=5.46,p=.065$, indicating that the environment did not have a significant effect on the number of pick failures.

A significant interaction effect was found between order and condition, $\chi^2(4)=13.81,p=.008$, suggesting that the effect of order on pick failures depends on the condition. The intercept was also significant, $\chi^2(1)=52.37,p<.001$, indicating a significant baseline level of pick failures.

Estimated marginal means for this model, given in \tabref{tab:failures}, were averaged over levels of order and environment. They are plotted along with the underlying observations in \figref{fig:failures}. Pairwise \textit{t} tests were conducted, with \textit{p} values adjusted using the Tukey method for comparing a family of 3 estimates to account for multiple comparisons.

\subsection{Place Failure Count}

Pick failure models incorporated fixed condition and random participant
effects.

\[
\text{PlaceFailureCount} \sim \text{Condition} + (1 | \text{Subject})
\]

A Type III Wald chi-square test was conducted to examine the effect of condition on the number of placement failures. The main effect of condition was statistically significant, $\chi^2(2)=8.20,p=.017$, indicating that the number of placement failures differed across the levels of condition. The intercept was not statistically significant, $\chi^2(1)=0.63,p=.427$, suggesting that the number of placement failures in the control condition (CON) was typically indistinguishable from zero.

Estimated marginal means for the place failure model are given in \tabref{tab:failures} and plotted along with the underlying observations in \figref{fig:failures}. \textit{p} values were adjusted using the Tukey method for comparing a family of 3 estimates to account for multiple comparisons.

\subsection{Workload}

Factors for condition order and block configuration (which of the three block scatterings, shown in \figref{fig:task-environments} was used for the trial) were considered, but did not significantly affect the model's outcome or fit, and are not included in the final analysis.

\[
\text{Workload} \sim \text{Condition} + (1 | \text{Subject})
\]

A Type III Wald chi-square test was conducted to examine the effect of condition on workload. The effect was statistically significant, $\chi^2(2)=15.27,p<.001$, indicating that workload differed across conditions. Estimated marginal means of the workload model are given in \tabref{tab:tlx}. Pairwise \textit{t} tests were conducted, and \textit{p} values were adjusted using the Tukey method for comparing a family of 3 estimates to account for multiple comparisons.

\subsection{Survival Analysis}

The Kaplan-Meier estimator was used to characterize participant's progression, with the resulting model shown in \figref{fig:survival-analysis}. While surviving longer is usually the desired observation in a survival analysis (e.g. when analyzing mortality data of patients receiving experimental medical interventions), our objective is for participants to complete the task more quickly. We invert the typical Y-axis ``survival'' rate and display completion (or ``mortality'') instead, so that the plot may still be read as higher-is-better.

\clearpage

\end{document}